\documentclass[11pt]{article}
\usepackage{acl2015}
\usepackage{times}
\usepackage{latexsym}
\usepackage{amsmath}
\usepackage{multirow}
\usepackage{url}
\usepackage[utf8]{inputenc}
\usepackage{graphicx}
\usepackage[encapsulated]{CJK}
\usepackage{ucs}
\usepackage{rotating}
\usepackage{algorithmic}
\usepackage{algorithm}
\usepackage{color}
\usepackage{amssymb}
\usepackage[usenames,dvipsnames,svgnames,table]{xcolor}

\newcommand{\ignore}[1]{}
\newcommand{\examp}[1]{\emph{#1}} 

\definecolor{MyGray}{rgb}{0.90,0.91,0.92}



\title{Finding Function in Form: Compositional Character Models for\\ Open Vocabulary Word Representation}
{\author{Wang Ling ~~~ Tiago Lu\'{\i}s ~~~ Lu\'{\i}s Marujo ~~~ Ram\'on Fernandez Astudillo\\ \textbf{Silvio Amir ~~~ Chris Dyer ~~~ Alan W Black ~~~ Isabel Trancoso} \\\\
  L$^2$F Spoken Systems Lab, INESC-ID, Lisbon, Portugal\\
  Language Technologies Institute, Carnegie Mellon University, Pittsburgh, PA, USA\\
  Instituto Superior T\'{e}cnico, Lisbon, Portugal\\
{\tt \{lingwang,lmarujo,cdyer,awb\}@cs.cmu.edu} \\ \tt \{ramon.astudillo,samir,tmcl,isabel.trancoso\}@inesc-id.pt\\}
}

\date{}

\begin{document}
\maketitle
\begin{CJK*}{UTF8}{gbsn}
\begin{abstract}

We introduce a model for constructing vector representations of words by composing characters using bidirectional LSTMs. Relative to traditional word representation models that have independent vectors for each word type, our model requires only a single vector per character type and a fixed set of parameters for the compositional model. Despite the compactness of this model and, more importantly, the arbitrary nature of the form--function relationship in language, our ``composed'' word representations yield state-of-the-art results in language modeling and part-of-speech tagging. Benefits over traditional baselines are particularly pronounced in morphologically rich languages (e.g., Turkish).
\end{abstract}

\section{Introduction}
Good representations of words are important for good generalization in natural language processing applications. Of central importance are vector space models that capture functional (i.e., semantic and syntactic) similarity in terms of geometric locality. However, when word vectors are learned---a practice that is becoming increasingly common---most models assume that each word type has its own vector representation that can vary independently of other model components. This paper argues that this independence assumption is inherently problematic, in particular in morphologically rich languages (e.g., Turkish). In such languages, a more reasonable assumption would be that orthographic (formal) similarity is evidence for functional similarity.

However, it is manifestly clear that similarity in form is neither a necessary nor sufficient condition for similarity in function: small orthographic differences may correspond to large semantic or syntactic differences (\emph{butter} vs. \emph{batter}), and large orthographic differences may obscure nearly perfect functional correspondence (\emph{rich} vs. \emph{affluent}).  Thus, any orthographically aware model must be able to capture \emph{non-compositional} effects in addition to more regular effects due to, e.g., morphological processes. To model the complex form--function relationship, we turn to long short-term memories (LSTMs), which are designed to be able to capture complex non-linear and non-local dynamics in sequences~\cite{Hochreiter:1997:LSM:1246443.1246450}. We use bidirectional LSTMs to ``read'' the character sequences that constitute each word and combine them into a vector representation of the word. This model assumes that each character type is associated with a vector, and the LSTM parameters encode both idiosyncratic lexical and regular morphological knowledge.

To evaluate our model, we use a vector-based model for part-of-speech (POS) tagging and for language modeling, and we report experiments on these tasks in several languages comparing to baselines that use more traditional, orthographically-unaware parameterizations. These experiments show: (i) our character-based model is able to generate similar representations for words that are semantically and syntactically similar, even for words are orthographically distant (e.g., \examp{October} and \examp{January}); our model achieves improvements over word lookup tables using only a fraction of the number of parameters in two tasks; (iii) our model obtains state-of-the-art performance on POS tagging (including establishing a new best performance in English); and (iv) performance improvements are especially dramatic in morphologically rich languages.

The paper is organized as follows:  Section~\ref{sec:c2v} presents our character-based model to generate word embeddings. Experiments on Language Modeling and POS tagging are described in Sections~\ref{sec:lang} and~\ref{sec:pos}. We present related work in Section~\ref{sec:relwork}; and we conclude in Section~\ref{sec:conclusions}.

\section{Word Vectors and Wordless Word Vectors}
\label{sec:c2v}
\ignore{Most NLP methods convert words into a sparse representation, where word types are treated independently. For instance, multinomial distributions that compute the probability of word $w$ being labelled as class $c$ are generally modelled with a table containing an a row for each word type and a column for each possible class. Similarly, in logistic regression a word is converted into ont-hot representation $\mathrm{onehot}(w)$, which is a vector with the size of the vocabulary $V$ and contains the value 1 in index $w$ and zero in all other indexes. As a softmax is performed on this vector, each position of the vector is given its own set of parameters. Thus, in both approaches, the number of parameters required to learn the model is $V\times C$, where $V$ and $C$ are the vocabulary and set of output classes. The obvious drawback of this representation is that if $V$ is very large (e.g., order of millions) it would be computationally expensive to store the parameters in memory. This is why, most unsupervised methods that are trained on large amounts of data, such as Brown Clustering, which requires a multinomial from word types to each cluster, prunes the vocabulary to a tractable size, leading to losses in terms of coverage. Additionally, the independence assumption leads to sparcity problems. As implied by the Zipf's law, only a relatively small number of word types will actually occur frequently enough for the models to generalize, while most word types will rarely occur leading to overfitting. 

\subsection{Word Lookup Tables}}
It is commonplace to represent words as vectors. In contrast to na\"{\i}ve models in which all word types in a vocabulary $V$ are equally different from each other, vector space models capture the intuition that words may be different or similar along a variety of dimensions. Learning vector representations of words by treating them as optimizable parameters in various kinds of language models has been found to be a remarkably effective means for generating vector representations that perform well in other tasks~\cite{collobert2011natural,kalchbrenner2013recurrent,liu2014recursive,chen2014fast}. Formally, such models define a matrix $\mathbf{P} \in \mathbb{R}^{d \times |V|}$, which contains $d$ parameters for each word in the vocabulary $V$. For a given word type $w\in V$, a column is selected by right-multiplying $\mathbf{P}$ by a one-hot vector of length $|V|$, which we write $\mathbf{1}_w$, that is zero in every dimension except for the element corresponding to $w$. Thus, $\mathbf{P}$ is often referred to as word lookup table and we shall denote by $\mathbf{e}^W_w \in \mathbb{R}^d$ the embedding obtained from a word lookup table for $w$ as $\mathbf{e}^W_w = \mathbf{P}\cdot \mathbf{1}_w$. This allows tasks with low amounts of annotated data to be trained jointly with other tasks with large amounts of data and leverage the similarities in these tasks. A common practice to this end is to initialize the word lookup table with the parameters trained on an unsupervised task. Some examples of these include the skip-$n$-gram and CBOW models of~\newcite{mikolov2013distributed}.

\subsection{Problem: Independent Parameters}
There are two practical problems with word lookup tables. Firstly, while they can be pre-trained with large amounts of data to learn semantic and syntactic similarities between words, each vector is independent. That is, even though models based on word lookup tables are often observed to learn that \examp{cats}, \examp{kings} and \examp{queens} exist in roughly the same linear correspondences to each other as \examp{cat}, \examp{king} and \examp{queen} do, the model does not represent the fact that adding an \examp{s} at the end of the word is evidence for this transformation. This means that word lookup tables cannot generate representations for previously unseen words, such as \examp{Frenchification}, even if the components, \examp{French} and \examp{-ification}, are observed in other contexts.

Second, even if copious data is available, it is impractical to actually store vectors for all word types. As each word type gets a set of parameters $d$, the total number of parameters is $d\times |V|$, where $|V|$ is the size of the vocabulary. Even in relatively morphological poor English, the number of word types tends to scale to the order of hundreds of thousands, and in noisier domains, such as online data, the number of word types raises considerably. For instance, in the English wikipedia dump with 60 million sentences, there are approximately 20 million different lowercased and tokenized word types, each of which would need its own vector. Intuitively, it is not sensible to use the same number of parameters for each word type.

Finally, it is important to remark that it is uncontroversial among cognitive scientists that our lexicon is structured into related forms---i.e., their parameters are not independent. The well-known ``past tense debate'' between connectionists and proponents of symbolic accounts concerns disagreements about how humans represent knowledge of inflectional processes (e.g., the formation of the English past tense), not whether such knowledge exists \cite{mw:1998}.


\subsection{Solution: Compositional Models}
Our solution to these problems is to construct a vector representation of a word by composing smaller pieces into a representation of the larger form. This idea has been explored in prior work by composing \emph{morphemes} into representations of words \cite{W13-3512,Botha2014,soricut:2015}. Morphemes are an ideal primitive for such a model since they are---by definition---the minimal meaning-bearing (or syntax-bearing) units of language. The drawback to such approaches is they depend on a morphological analyzer.

In contrast, we would like to compose representations of \emph{characters} into representations of words. However, the relationship between words forms and their meanings is non-trivial \cite{saussure:2013}. While some compositional relationships exist, e.g., morphological processes such as adding \emph{-ing} or \emph{-ly} to a stem have relatively regular effects, many words with lexical similarities convey different meanings, such as, the word pairs \examp{lesson}$\iff$\examp{lessen} and \examp{coarse}$\iff$\examp{course}.

\section{C2W Model}

\begin{figure}[ht]
\begin{center}
\centerline{\includegraphics[width=\columnwidth]{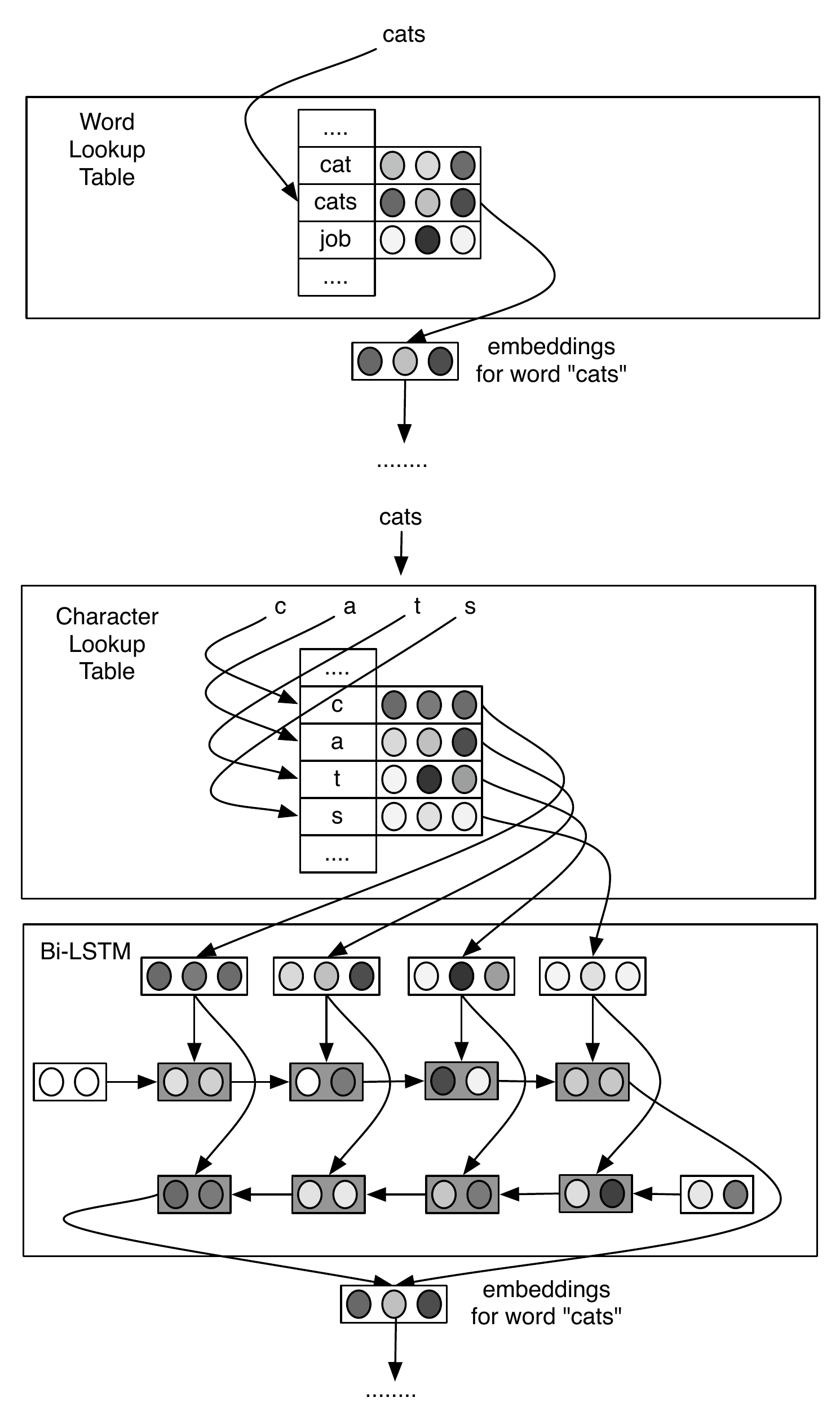}}
\caption{Illustration of the word lookup tables (top) and the lexical Composition Model (bottom). Square boxes represent vectors of neuron activations. Shaded boxes indicate that a non-linearity. }
\label{model}
\end{center}
\end{figure}

Our compositional character to word (C2W) model is based on bidirectional LSTMs~\cite{journals/nn/GravesS05}, which are able to learn complex non-local dependencies in sequence models. An illustration is shown in Figure~\ref{model}. The input of the C2W model (illustrated on bottom) is a single word type $w$, and we wish to obtain is a $d$-dimensional vector used to represent $w$. This model shares the same input and output of a word lookup table (illustrated on top), allowing it to easily replace then in any network.

As input, we define an alphabet of characters $C$. For English, this vocabulary would contain an entry for each uppercase and lowercase letter as well as numbers and punctuation. The input word $w$ is decomposed into a sequence of characters $c_1,\ldots,c_m$, where $m$ is the length of $w$. Each $c_i$ is defined as a one hot vector $\mathbf{1}_{c_i}$, with one on the index of $c_i$ in vocabulary $M$. We define a projection layer $\mathbf{P}_C \in \mathbb{R}^{d_C \times |C|}$, where $d_C$ is the number of parameters for each character in the character set $C$. This of course just a character lookup table, and is used to capture similarities between characters in a language (e.g., vowels \emph{vs}. consonants). Thus, we write the projection of each input character $c_i$ as $\mathbf{e}_{c_i}=\mathbf{P}_C \cdot \mathbf{1}_{c_i}$.

Given the input vectors $\mathbf{x}_1,\ldots,\mathbf{x}_m$, a LSTM computes the state sequence $\mathbf{h}_1,\ldots,\mathbf{h}_{m+1}$ by iteratively applying the following updates:

\begin{align*}
\mathbf{i}_t &= \sigma(\mathbf{W}_{ix}\mathbf{x}_t + \mathbf{W}_{ih}\mathbf{h}_{t-1} + \mathbf{W}_{ic}\mathbf{c}_{t-1} + \mathbf{b}_i) \\
\mathbf{f}_t &= \sigma(\mathbf{W}_{fx}\mathbf{x}_t + \mathbf{W}_{fh}\mathbf{h}_{t-1} + \mathbf{W}_{fc}\mathbf{c}_{t-1} + \mathbf{b}_f) \\
\mathbf{c}_t &= \mathbf{f}_t \odot \mathbf{c}_{t-1} + \\
& {}\ \ \qquad \mathbf{i}_t \odot \tanh(\mathbf{W}_{cx}\mathbf{x}_t +  \mathbf{W}_{ch}\mathbf{h}_{t-1} + \mathbf{b}_c)\\
\mathbf{o}_t &= \sigma(\mathbf{W}_{ox}\mathbf{x}_t + \mathbf{W}_{oh}\mathbf{h}_{t-1} + \mathbf{W}_{oc}\mathbf{c}_{t} + \mathbf{b}_o) \\
\mathbf{h}_t &= \mathbf{o}_t \odot \tanh(\mathbf{c}_t),
\end{align*}
where $\sigma$ is the component-wise logistic sigmoid function, and $\odot$ is the component-wise (Hadamard) product. LSTMs define an extra cell memory $\mathbf{c}_t$, which is combined linearly at each timestamp $t$. The information that is propagated from $\mathbf{c}_{t-1}$ to $\mathbf{c}_{t}$ is controlled by the three gates $\mathbf{i}_t$, $\mathbf{f}_t$,  and $\mathbf{o}_t$, which determine the what to include from the input $\mathbf{x}_t$, the what to forget from $\mathbf{c}_{t-1}$ and what is relevant to the current state $\mathbf{h}_t$. We write $\mathcal{W}$ to refer to all parameters the LSTM ($\mathbf{W}_{ix}$, $\mathbf{W}_{fx}$, $\mathbf{b}_f$, \ldots). Thus, given a sequence of character representations $\mathbf{e}^C_{c_1},\ldots,\mathbf{e}^C_{c_m}$ as input, the forward LSTM, yields the state sequence $\mathbf{s}^f_0,\ldots,\mathbf{s}^f_m$, while the backward LSTM receives as input the reverse sequence, and yields states $\mathbf{s}^b_m,\ldots,\mathbf{s}^b_0$. Both LSTMs use a different set of parameters $\mathcal{W}^f$ and $\mathcal{W}^b$. The representation of the word $w$ is obtained by combining the forward and backward states:
\begin{align*}
\mathbf{e}^C_w = \mathbf{D}^f \mathbf{s}^f_m + \mathbf{D}^b \mathbf{s}^b_0 + \mathbf{b}_{d},
\end{align*}
where $\mathbf{D}^f$, $\mathbf{D}^b$ and $\mathbf{b}_{d}$ are parameters that determine how the states are combined. 

\paragraph{Caching for Efficiency.} Relative to $\mathbf{e}^W_w$, computing $\mathbf{e}^C_w$ is computational expensive, as it requires two LSTMs traversals of length $m$. However, $\mathbf{e}^C_w$ only depends on the character sequence of that word, which means that unless the parameters are updated, it is possible to cache the value of $\mathbf{e}^C_w$ for each different $w$'s that will be used repeatedly. Thus, the model can keep a list of the most frequently occurring word types in memory and run the compositional model only for rare words. Obviously, caching all words would yield the same performance as using a word lookup table $\mathbf{e}^W_w$, but also using the same amount of memory. Consequently, the number of word types used in cache can be adjusted to satisfy memory vs. performance requirements of a particular application.

At training time, when parameters are changing, repeated words within the same batch only need to be computed once, and the gradient at the output can be accumulated within the batch so that only one update needs to be done per word type. For this reason, it is preferable to define larger batches.

\section{Experiments: Language Modeling}
\label{sec:lang}

Our proposed model is similar to models used to compute composed representations of sentences from words ~\cite{cho-EtAl:2014:EMNLP2014,DBLP:journals/corr/LiJH15}. However, the relationship between the meanings of individual words and the composite meaning of a phrase or sentence is arguably more regular than the relationship of representations of characters and the meaning of a word. Is our model capable of learning such an irregular relationship? We now explore this question empirically. 

Language modeling is a task with many applications in NLP.  An effective LM requires syntactic aspects of language to be modeled, such as word orderings (e.g., ``John is smart" \emph{vs}. ``John smart is"), but also semantic aspects (e.g., ``John ate fish" \emph{vs}. ``fish ate John"). Thus, if our C2W model only captures regular aspects of words, such as, prefixes and suffixes, the model will yield worse results compared to word lookup tables.

\subsection{Language Model}
Language modeling amounts to learning a function that computes the log probability, $\log p(\boldsymbol{w})$, of a sentence $\boldsymbol{w}=(w_1,\ldots,w_n)$. This quantity can be decomposed according to the chain rule into the sum of the conditional log probabilities $\sum_{i=1}^{n} \log p(w_i \mid w_1,\ldots,w_{i-1})$. Our language model computes $\log p(w_i \mid w_1,\ldots,w_{i-1})$ by composing representations of words $w_1,\ldots,w_{i-1}$ using an recurrent LSTM model~\cite{mikolov2010recurrent,sundermeyer12:lstm}.

\begin{figure}[ht]
\begin{center}
\centerline{\includegraphics[width=\columnwidth]{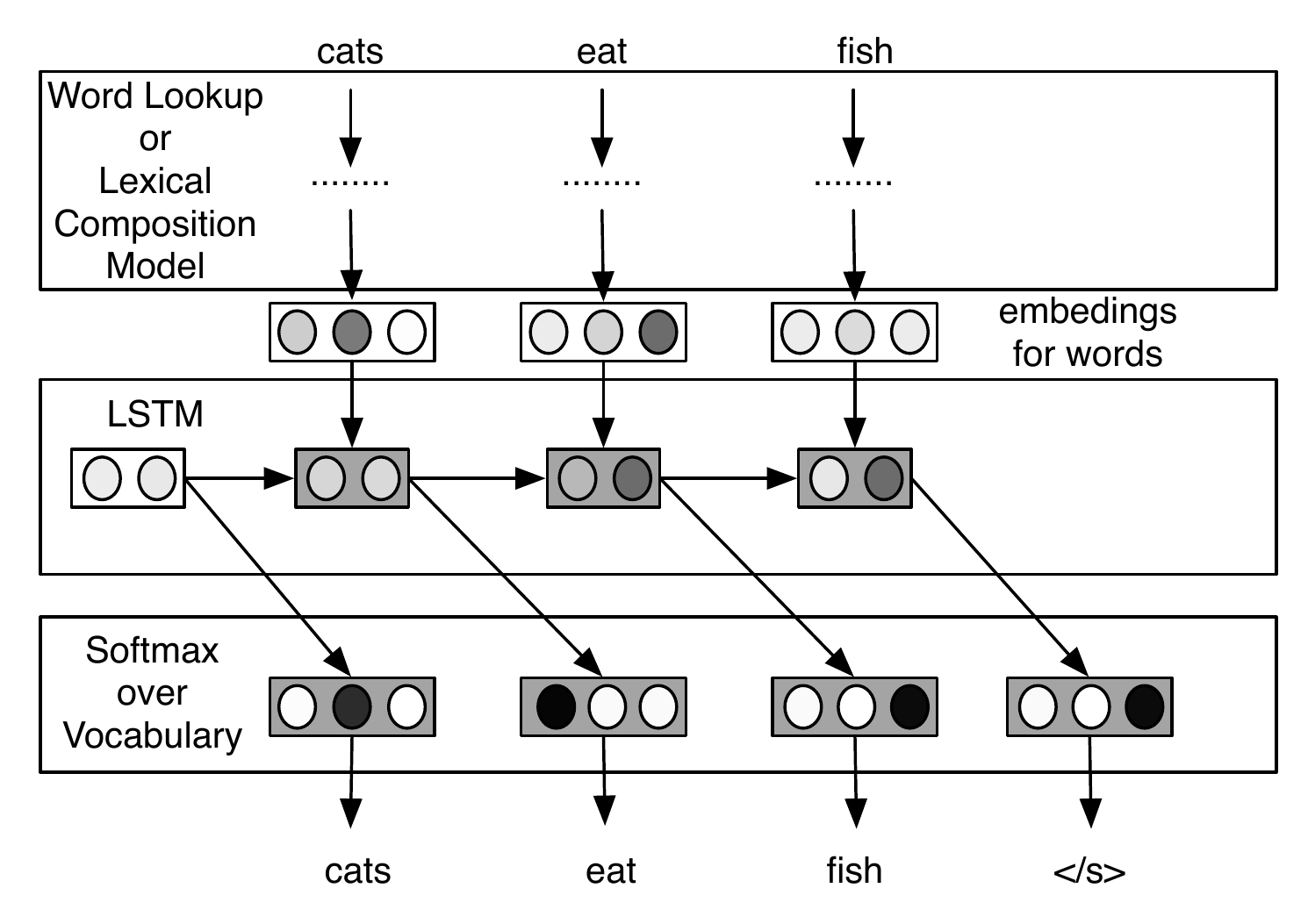}}
\caption{Illustration of our neural network for Language Modeling.}
\label{lm}
\end{center}
\end{figure}

The model is illustrated in Figure~\ref{lm}, where we observe on the first level that each word $w_i$ is projected into their word representations. This can be done by using word lookup tables $\mathbf{e}^W_{w_i}$, in which case, we will have a regular recurrent language model. To use our C2W model, we can simply replace the word lookup table with the model $f(w_i)=\mathbf{e}^C_{w_i}$. Each LSTM block $\mathbf{s}_i$, is used to predict word $w_{i+1}$. This is performed by projecting the $s_i$ into a vector of size of the vocabulary $V$ and performing a softmax.

The softmax is still simply a $d\times V$ table, which encodes the likelihood of every word type in a given context, which is a closed-vocabulary model. Thus, at test time out-of-vocabulary (OOV) words cannot be addressed. A strategy that is generally applied is to prune the vocabulary $V$ by replacing word types with lower frequencies as an OOV token. At test time, the probability of words not in vocabulary is estimated as the OOV token. Thus, depending on the number of word types that are pruned, the global perplexities may decrease, since there are fewer outcomes in the softmax, which makes the absolute value of perplexity not informative when comparing models of different vocabulary sizes. Yet, the relative perplexity between different models indicates which models can better predict words based on their contexts. 

To address OOV words in the baseline setup, these are replaced by an unknown token, and also associated with a set of embeddings. During training, word types that occur once are replaced with the unknown token stochastically with 0.5 probability. The same process is applied at the character level for the C2W model.

\subsection{Experiments}
\paragraph{Datasets}
\ignore{Languages can be characterized by the morphological processes they use as being either analytic or synthetic. Analytic languages, such as English, contain relatively little inflection, and rely on word order and auxiliary words to express grammatical relations. Synthetic languages can be further divided into fusional and agglutinative languages. While both allow words to be inflected, . }
We look at the language model performance on English, Portuguese, Catalan, German and Turkish, which have a broad range of morphological typologies. While all these languages contain inflections, in agglutinative languages affixes tend to be unchanged, while in fusional languages they are not. For each language, Wikipedia articles were randomly extracted until 1 million words are obtained and these were used for training. For development and testing, we extracted an additional set of 20,000 words.

\paragraph{Setup}
We define the size of the word representation $d$ to 50. In the C2W model requires setting the dimensionality of characters $d_C$ and current states $d_{CS}$. We set $d_C=50$ and $d_{CS}=150$. Each LSTM state used in the language model sequence $s_i$ is set to 150 for both states and cell memories. Training is performed with mini-batch gradient descent with 100 sentences. The learning rate and momentum were set to 0.2 and 0.95. The softmax over words is always performed on lowercased words. \ignore{the softmax operation can be computationally expensive to compute over large vocabularies 100K words. While there are more efficient methods to approximate the softmax function~\cite{Mnih12afast}, as our work is focused on learning better representations, we decided to use a simpler approach.} We restrict the output vocabulary to the most frequent 5000 words. Remaining word types will be replaced by an unknown token, which must also be predicted. The word representation layer is still performed over all word types (i.e., completely open vocabulary). When using word lookup tables, the input words are also lowercased, as this setup produces the best results. In the C2W, case information is preserved. 

Evaluation is performed by computing the perplexities over the test data, and the parameters that yield the highest perplexity over the development data are used.

\paragraph{Perplexities}
Perplexities over the testset are reported on Table~\ref{lang}. From these results, we can see that in general, it is clear that C2W always outperforms word lookup tables (row ``Word"), and that improvements are especially pronounced in Turkish, which is a highly morphological language, where word meanings differ radically depending on the suffixes used (\examp{evde} $\to$ \examp{in the house} \emph{vs.} \examp{evden} $\to$ \examp{from the house}). \ignore{ For instance, the Turkish word for \examp{ev}~(English \examp{house}), attached with the suffix \examp{-de}, changes its meaning to \examp{in the house}, while adding the suffix \examp{-den} would mean \examp{from the house}. Thus, while the C2W model can learn that the suffix \examp{-de} means inside the stem of the word, word lookup tables would need to have examples for every word with the suffix \examp{-de} to learn good vector representations. Furthermore, as it is not uncommon for multiple suffixes to be added to a word, suffix features would not suffice to learn this morphological process. Improvements on other languages are not as significant, but consistent. For instance, in Portuguese, gender is encoded into nouns as suffixes, for instance, a male cat is spelled as \examp{gato}, a female one as \examp{gata}. Thus, a word lookup table would require all nouns to occur in both forms, while the C2W model can also generalize this process.}

\paragraph{Number of Parameters}
As for the number of parameters (illustrated for block ``\#Parameters"), the number of parameters in word lookup tables is $V\times d$. If a language contains 80,000 word types (a conservative estimate in morphologically rich languages), 4 million parameters would be necessary.  On the other hand, the compositional model consists of 8 matrices of dimensions $d_{CS} \times d_C+2d_{CS}$. Additionally, there is also the matrix that combines the forward and backward states of size $d \times 2d_{CS}$. Thus, the number of parameters is roughly 150,000 parameters---substantially fewer. This model also needs a character lookup table with $d_C$ parameters for each entry. For English, there are 618 characters, for an additional 30,900 parameters. So the total number of parameters for English is roughly 180,000 parameters (2 to 3 parameters per word type), which is an order of magnitude lower than word lookup tables.

\paragraph{Performance}
As for efficiency, both representations can label sentences at a rate of approximately 300 words per second during training. While this is surprising, due to the fact that the C2W model requires a composition over characters, the main bottleneck of the system is the softmax over the vocabulary. Furthermore, caching is used to avoid composing the same word type twice in the same batch. This shows that the C2W model, is relatively fast compared operations such as a softmax.

\begin{table}
\begin{center}
\scalebox{0.85}{
\begin{tabular}{l|c|c|c|c|c}
 & \multicolumn{3}{c}{Fusional} & \multicolumn{2}{|c}{Agglutinative}\\
\hline
Perplexity & EN & PT & CA & DE & TR\\
\hline
5-gram KN & 70.72 & 58.73 & 39.83 & 59.07 & 52.87\\
Word & 59.38 & 46.17 & 35.34 & 43.02 & 44.01 \\
C2W & \textbf{57.39} & \textbf{40.92} & \textbf{34.92} & \textbf{41.94} & \textbf{32.88} \\
\hline
\#Parameters & & & & & \\
\hline
Word & 4.3M & 4.2M & 4.3M & 6.3M & 5.7M \\
C2W & \textbf{180K} & \textbf{178K} & \textbf{182K} & \textbf{183K} & \textbf{174K} \\
\end{tabular}
}
\end{center}
\caption{\label{lang} Language Modeling Results}
\end{table}

\ignore{\paragraph{Word Representations}

While the C2W model does not keep word vectors explicitly, we can still look at word similarity tasks by generating words representations from a list of words. Thus, we generate continuous representations for all words in our English Wikipedia dataset to evaluate what the model is learning. Table~\ref{top-10} shows the 10-closest words ranked by cosine distance in vector space for the terms ``increased" and ``John" using the C2W model. We observe from this list that while the C2W model does prefer to group words with similar suffixes, it can effectively group words that are semantically similar. For word ``increased,'' we can see that the closest words are other verbs that have similar suffixes but also similar meanings. For the proper noun ``John", the list contains other names that are lexically divergent.}

\paragraph{Representations of (nonce) words}
While is is promising that the model is not simply learning lexical features, what is most interesting is that the model can propose embeddings for nonce words, in stark contrast to the situation observed with lookup table models. We show the 5-most-similar in-vocabulary words (measured with cosine similarity) as computed by our character model on two in-vocabulary words and two nonce words\footnote{software submitted as supplementary material}.\ignore{
For instance, if we add to the word \examp{phd} the suffix \examp{-ing}, it will respond with a top-10 list comprising of verbs and adjectives as these are the most likely word functions for the invented word, based on the suffix. If we take a proper name \examp{Noah} and add \examp{shire}, we obtain a list with the names of many locations, such as Petersburg and Zurich. This is because, many locations are composed of peoples names followed by suffixes, such as \examp{burg}, \examp{shire} and \examp{land}, and these regularities are captured by our model.}This makes our model generalize significantly better than lookup tables that generally use unknown tokens for OOV words. Furthermore, this ability to generalize is much more similar to that of human beings, who are able to infer meanings for new words based on its form.

\begin{table}
\begin{center}
\scalebox{0.80}{
\begin{tabular}{c|c||c|c}
\emph{increased}& \emph{John} & \emph{Noahshire}  & \emph{phding} \\
\hline
reduced & Richard & Nottinghamshire & mixing\\
improved & George & Bucharest & modelling \\
expected & James & Saxony & styling \\
decreased & Robert & Johannesburg & blaming \\
targeted & Edward & Gloucestershire & christening \\
\hline
\end{tabular}
}
\end{center}
\caption{\label{top-5} Most-similar in-vocabular words under the C2W model; the two query words on the left are in the training vocabulary, those on the right are nonce (invented) words.}
\end{table} 

\section{Experiments: Part-of-speech Tagging}
\label{sec:pos}

As a second illustration of the utility of our model, we turn to POS tagging. As morphology is a strong indicator for syntax in many languages, a much effort has been spent engineering features~\cite{Nakagawa01unknownword,D13-1032}. We now show that some of these features can be learnt automatically using our model. 

\subsection{Bi-LSTM Tagging Model}

\begin{figure}[ht]
\begin{center}
\centerline{\includegraphics[width=\columnwidth]{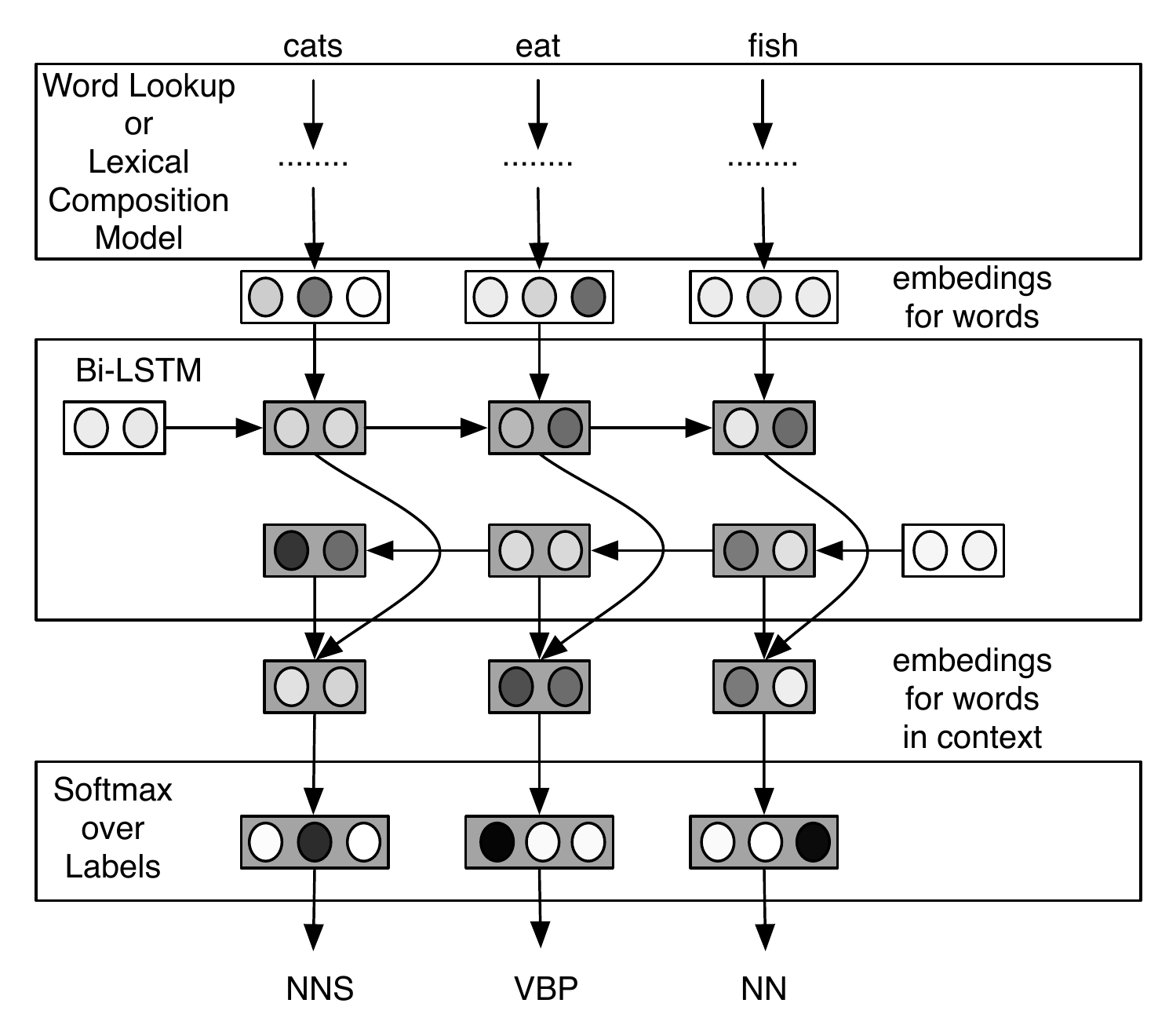}}
\caption{Illustration of our neural network for POS tagging.}
\label{bow}
\end{center}
\end{figure}

Our tagging model is likewise novel, but very straightforward. It builds a Bi-LSTM over words as illustrated in Figure~\ref{bow}. The input of the model is a sequence of features $f(w_1),\ldots,f(w_n)$. Once again, word vectors can either be generated using the C2W model $f(w_i)=\mathbf{e}^C_{w_i}$, or word lookup tables $f(w_i)=\mathbf{e}^W_{w_i}$. We also test the usage of hand-engineered features, in which case  $f_1(w_i),\ldots,f_n(w_i)$. Then, the sequential features $f(w_1),\ldots,f(w_n)$ are fed into a bidirectional LSTM model, obtaining the forward states $\mathbf{s}^f_0,\ldots,\mathbf{s}^f_n$ and the backward states $\mathbf{s}^b_{N+1},\ldots,\mathbf{s}^b_{0}$. Thus, state $\mathbf{s}^f_i$ contains the information of all words from $0$ to $i$ and $\mathbf{s}^b_i$ from $n$ to $i$. The forward and backward states are combined, for each index from $1$ to $n$, as follows:
\begin{align*}
\mathbf{l}_i = \tanh(\mathbf{L}^f \mathbf{s}^f_i + \mathbf{L}^b \mathbf{s}^b_i + \mathbf{b}_{l}),
\end{align*}
where $\mathbf{L}^f$, $\mathbf{L}^b$ and $\mathbf{b}_{l}$ are parameters defining how the forward and backward states are combined. The size of the forward $\mathbf{s}^f$ and backward states $\mathbf{s}^b$ and the combined state $\mathbf{l}$ are hyperparameters of the model, denoted as $d^f_{WS}$, $d^b_{WS}$ and $d_{WS}$, respectively. Finally, the output labels for index $i$ are obtained as a softmax over the POS tagset, by projecting the combined state $\mathbf{l}_i$.

\label{sec:posmodel}

\subsection{Experiments}
\paragraph{Datasets}
For English, we conduct experiments on the Wall Street Journal of the Penn Treebank dataset~\cite{Marcus:1993:BLA:972470.972475}, using the standard splits (sections 1--18 for train, 19--21 for tuning and 22--24 for testing). We also perform tests on 4 other languages, which we obtained from the CoNLL shared tasks~\cite{marti2007cess,Brants02thetiger,pt,tr}. While the PTB dataset provides standard train, tuning and test splits, there are no tuning sets in the datasets in other languages, so we withdraw the last 100 sentences from the training dataset and use them for tuning. 

\paragraph{Setup}
The POS model requires two sets of hyperparameters. Firstly, words must be converted into continuous representations and the same hyperparametrization as in language modeling~(Section~\ref{sec:lang}) is used. Secondly, words representations are combined to encode context. Our POS tagger has three hyperparameters $d^f_{WS}$, $d^b_{WS}$ and $d_{WS}$, which correspond to the sizes of LSTM states, and are all set to 50. As for the learning algorithm, use the same setup (learning rate, momentum and mini-batch sizes) as used in language modeling.

Once again, we replace OOV words with an unknown token, in the setup that uses word lookup tables, and the same with OOV characters in the C2W model. In setups using pre-trained word embeddings, we consider a word an OOV if it was not seen in the labelled training data as well as in the unlabeled data used for pre-training.

\paragraph{Compositional Model Comparison}
A comparison of different recurrent neural networks for the C2W model is presented in Table~\ref{wsjtest}. We used our proposed tagger tagger in all experiments and results are reported for the English Penn Treebank. Results on label accuracy test set is shown in the column ``acc". The number of parameters in the word composition model is shown in the column ``parameters". Finally, the number of words processed at test time per second are shown in column ``words/sec".

We observe that approaches using RNN yield worse results than their LSTM counterparts with a difference of approximately 2\%.  This suggests that while regular RNNs can learn shorter character sequence dependencies, they are not ideal to learn longer dependencies. LSTMs, on the other hand, seem to effectively obtain relatively higher results, on par with using word look up tables (row ``Word Lookup"), even when using forward (row ``Forward LSTM") and backward (row ``Backward LSTM") LSTMs individually. The best results are obtained using the bidirectional LSTM (row ``Bi-LSTM"), which achieves an accuracy of 97.29\% on the test set, surpassing the word lookup table. 

There are approximately 40k lowercased word types in the training data in the PTB dataset. Thus, a word lookup table with 50 dimensions per type contains approximately 2 million parameters. In the C2W models, the number of characters types (including uppercase and lowercase) is approximately 80. Thus, the character look up table consists of only 4k parameters, which is negligible compared to the number of parameters in the compositional model, which is once again 150k parameters. One could argue that results in the Bi-LSTM model are higher than those achieved by other models as it contains more parameters, so we set the state size $d_{CS}=50$ (row ``Bi-LSTM $d_{CS}=50$") and obtained similar results.

In terms of computational speed, we can observe that there is a more significant slowdown when applying the C2W models compared to language modeling. This is because there is no longer a softmax over the whole word vocabulary as the main bottleneck of the network. However, we can observe that while the Bi-LSTM system is 3 times slower, it is does not significantly hurt the performance of the system.

\begin{table}
\begin{center}
\scalebox{0.85}{
\begin{tabular}{c|c|c|c}
 & acc & parameters & words/sec\\
\hline
Word Lookup & 96.97 & 2000k & 6K\\
\hline
Forward RNN &  95.66 & 17.5k & 4K\\
Backward RNN & 95.52 & 17.5k & 4K\\
Bi-RNN & 95.93 & 40k & 3K\\
Forward LSTM & 97.12 & 80k & 3K\\
Backward LSTM & 97.08 & 80k & 3K\\
Bi-LSTM $d_{CS}=50$ & 97.22 & 70k & 3K\\
Bi-LSTM & \textbf{97.36} & 150k & 2K\\
\hline

\end{tabular}
}
\end{center}
\caption{\label{wsjtest} POS accuracy results for the English PTB using word representation models.}
\end{table}

\paragraph{Results on Multiple Languages}
Results on 5 languages are shown in Table~\ref{lang}. In general, we can observe that the model using word lookup tables~(row ``Word") performs consistently worse than the C2W model~(row ``C2W"). We also compare our results with Stanford's POS tagger, with the default set of features, found in Table~\ref{lang}. Results using these tagger are comparable or better than state-of-the-art systems. We can observe that in most cases we can slightly outperform the scores obtained using their tagger. This is a promising result, considering that we use the same training data and do not handcraft any features.  Furthermore, we can observe that for Turkish, our results are significantly higher ($>$4\%). 

\begin{table}
\begin{center}
\scalebox{0.85}{
\begin{tabular}{l|c|c|c|c|c}
System & \multicolumn{3}{c}{Fusional} & \multicolumn{2}{|c}{Agglutinative}\\
\hline
 & EN & PT & CA & DE & TR\\
\hline
Word & 96.97 & 95.67 & 98.09 & 97.51 & 83.43 \\
C2W & \textbf{97.36} & 97.47 & \textbf{98.92} & \textbf{98.08} & \textbf{91.59} \\
Stanford & 97.32 & \textbf{97.54} & 98.76 & 97.92 & 87.31\\
\hline
\end{tabular}
}
\end{center}
\caption{\label{lang} POS accuracies on different languages}
\end{table}

\paragraph{Comparison with Benchmarks}

Most state-of-the-art POS tagging systems are obtained by either learning or handcrafting good lexical features~\cite{Manning:2011:PT9:1964799.1964816,DBLP:journals/corr/Sun14c} or using additional raw data to learn features in an unsupervised fashion. Generally, optimal results are obtained by performing both. Table~\ref{wsjstate} shows the current Benchmarks in this task for the English PTB. Accuracies on the test set is reported on column ``acc". Columns ``feat" and ``data" define whether hand-crafted features are used and whether additional data was used. We can see that even without feature engineering or unsupervised pretraining, our C2W model (row ``C2W") is on par with the current state-of-the-art system (row ``structReg"). However, if we add hand-crafted features, we can obtain further improvements on this dataset (row ``C2W + features"). 

However, there are many words that do not contain morphological cues to their part-of-speech. For instance, the word \emph{snake} does not contain any morphological cues that determine its tag. In these cases, if they are not found labelled in the training data, the model would be dependent on context to determine their tags, which could lead to errors in ambiguous contexts. Unsupervised training methods such as the Skip-$n$-gram model~\cite{mikolov2013distributed} can be used to pretrain the word representations on unannotated corpora. If such pretraining places \emph{cat}, \emph{dog} and \emph{snake} near each other in vector space, and the supervised POS data contains evidence that \emph{cat} and \emph{dog} are nouns, our model will be likely to label \emph{snake} with the same tag.

We train embeddings using English wikipedia with the dataset used in~\cite{Ling:2015:naacl}, and the Structured Skip-$n$-gram model. Results using pre-trained word lookup tables and the C2W with the pre-trained word lookup tables as additional parameters are shown in rows ``word(sskip)" and ``C2W + word(sskip)". We can observe that both systems can obtain improvements over their random initializations (rows ``word" and (C2W)). 

Finally, we also found that when using the C2W model in conjunction pre-trained word embeddings, that adding a non-linearity to the representations extracted from the C2W model $\mathbf{e}^C_w$ improves the results over using a simple linear transformation (row ``C2W(tanh)$+$word~(sskip)"). This setup, obtains 0.28 points over the current state-of-the-art system(row ``SCCN").

A similar model that a convolutional model to learn additional representations for words~\cite{icml2014c2_santos14}~(row ``CNN~
\cite{icml2014c2_santos14}"). However, results are not directly comparable as a different set of embeddings is used to initialize the word lookup table.


\begin{table}
\begin{center}
\scalebox{0.85}{
\begin{tabular}{l|c|c|c}
& feat & data & acc\\
\hline
word & no & no & 96.70 \\
C2W & no & no & \textbf{97.36} \\
\hline
word$+$features & yes & no & 97.34 \\
C2W$+$features & yes & no & \textbf{97.57} \\
Stanford 2.0~\cite{Manning:2011:PT9:1964799.1964816} & yes & no & 97.32 \\
structReg~\cite{DBLP:journals/corr/Sun14c} & yes & no & 97.36 \\
\hline
word~(sskip) & no & yes & 97.42 \\
C2W$+$word~(sskip) & no & yes & 97.54 \\
C2W(tanh)$+$word~(sskip) & no & yes & \textbf{97.78} \\
CNN~\cite{icml2014c2_santos14} & no & yes & 97.32 \\
Mor\v{c}e~\cite{Spoustova:2009:STA:1609067.1609152} & yes & yes & 97.44 \\
SCCN~\cite{Sogaard:2011:SCN:2002736.2002748} & yes & yes & 97.50 \\
\end{tabular}
}
\end{center}
\caption{\label{wsjstate} POS accuracy result comparison with state-of-the-art systems for the English PTB.}
\end{table}

\subsection{Discussion}

It is important to refer here that these results do not imply that our model always outperforms existing benchmarks, in fact in most experiments, results are typically fairly similar to existing systems. Even in Turkish, using morphological analysers in order to extract additional features could also accomplish similar results. The goal of our work is not to overcome existing benchmarks, but show that much of the feature engineering done in the benchmarks can be learnt automatically from the task specific data. More importantly, we wish to show large dimensionality word look tables can be compacted into a lookup table using characters and a compositional model allowing the model scale better with the size of the training data. This is a desirable property of the model as data becomes more abundant in many NLP tasks.

\section{Related Work}
\label{sec:relwork}

Our work, which learns representations without relying on word lookup tables has not been explored to our knowledge. In essence, our model attempts to learn lexical features automatically while compacting the model by reducing the redundancy found in word lookup tables. Individually, these problems have been the focus of research in many areas.

Lexical information has been used to augment word lookup tables. Word representation learning can be thought of as a process that takes a string as input representing a word and outputs a set of values that represent a word in vector space. Using word lookup tables is one possible approach to accomplish this. Many methods have been used to augment this model to learn lexical features with an additional model that is jointly maximized with the word lookup table. This is generally accomplished by either performing a component-wise addition of the embeddings produced by word lookup tables~\cite{chenjoint}, and that generated by the additional lexical model, or simply concatenating both representations~\cite{icml2014c2_santos14}. Many models have been proposed, the work in~\cite{collobert2011natural} refers that additional features sets $F_i$ can be added to the one-hot representation and multiple lookup tables $\mathbf{I}_{F_i}$ can be learnt to project each of the feature sets to the same low-dimensional vector $\mathbf{e}^W_w$. For instance, the work in~\cite{Botha2014} shows that using morphological analyzers to generate morphological features, such as stems, prefixes and suffixes can be used to learn better representations for words. A problem with this approach is the fact that the model can only learn from what has been defined as feature sets. The models proposed in~\cite{icml2014c2_santos14,chenjoint} allow the model to arbitrary extract meaningful lexical features from words by defining compositional models over characters. The work in~\cite{chenjoint} defines a simple compositional model by summing over all characters in a given word, while the work in ~\cite{icml2014c2_santos14} defines a convolutional network, which combines windows of characters and a max-pooling layer to find important morphological features. The main drawback of these methods is that character order is often neglected, that is, when summing over all character embeddings, words such as \examp{dog} and \examp{god} would have the same representation according to the lexical model. Convolutional model are less susceptible to these problems as they combine windows of characters at each convolution, where the order within the window is preserved. However, the order between extracted windows is not, so the problem still persists for longer words, such as those found in agglutinative languages. Yet, these approaches work in conjunction with a word lookup table, as they compensate for this inability. Aside from neural approaches, character-based models have been applied to address multiple lexically oriented tasks, such as transliteration~\cite{kang2000automatic} and twitter normalization~\cite{xu-ritter-grishman:2013:BUCC,ling-EtAl:2013:EMNLP}.

Compacting models has been a focus of research in tasks, such as language modeling and machine translation, as extremely large models can be built with the large amounts of training data that are available in these tasks. In language modeling, it is frequent to prune higher order n-grams that do not encode any additional information~\cite{Seymore96scalablebackoff,Stolcke98entropy-basedpruning,Moore:2009:LMS:1699571.1699610}. The same be applied in machine translation~\cite{ling2012,rzens2012} by removing longer translation pairs that can be replicated using smaller ones. In essence our model learns regularities at the subword level that can be leveraged for building more compact word representations.

Finally, our work has been applied to dependency parsing and found similar improvements over word models in morphologically rich languages~\cite{ling:2015}.

\section{Conclusion}
\label{sec:conclusions}

We propose a C2W model that builds word embeddings for words without an explicit word lookup table. Thus, it benefits from being sensitive to lexical aspects within words, as it takes characters as atomic units to derive the embeddings for the word. On POS tagging, our models using characters alone can still achieve comparable or better results than state-of-the-art systems, without the need to manually engineer such lexical features. Although both language modeling and POS tagging both benefit strongly from morphological cues, the success of our models in languages with impoverished morphological cues shows that it is able to learn non-compositional aspects of how letters fit together. 

The code for the C2W model and our language model and POS tagger implementations is available from \url{https://github.com/wlin12/JNN}.

\end{CJK*} 
\section*{Acknowledgements}
%
The PhD thesis of Wang Ling is supported by FCT grant SFRH/BD/51157/2010.
This research was supported in part by the U.S.~Army Research Laboratory, the U.S.~Army Research Office
under contract/grant number W911NF-10-1-0533 and NSF IIS-1054319 and FCT through the plurianual contract UID/CEC/50021/2013 and grant number SFRH/BPD/68428/2010. 
\bibliographystyle{acl}
\bibliography{paper}

\end{document}